\documentclass[journal]{IEEEtran}

\ifCLASSINFOpdf
\else
   \usepackage[dvips]{graphicx}
\fi
\usepackage{url}
\usepackage[
    colorlinks = true,
    urlcolor  = blue,
]{hyperref}

\usepackage{graphicx}
\usepackage[%linesnumbered
,ruled]{algorithm2e}
\usepackage{amssymb}
\usepackage{mathtools}
\usepackage{comment}
\usepackage{xcolor}
\usepackage{booktabs}

\begin{document}

\title{Soft Gradient Boosting with Learnable Feature Transforms for Sequential Regression}

\author{Huseyin Karaca, Suleyman Serdar Kozat, \IEEEmembership{Senior Member, IEEE}
\thanks{Manuscript submitted to IEEE Signal Processing Letters. Last update at \today.
Corresponding author: Huseyin Karaca.}
\thanks{The authors are with the Department of Electrical and Electronics Engineering, Bilkent University, 06800 Ankara, Turkey (e-mail: huseyin.karaca@bilkent.edu.tr; kozat@ee.bilkent.edu.tr).}
\thanks{This work is partly supported by TUBİTAK (The Scientific and Technological Research Council of Turkey).}% under grant 2210-A.}
%\thanks{This paragraph of the first footnote  contain the date on which you submitted your paper for review. It  also contain support information, including sponsor and financial support acknowledgment. For example, ``This work was supported in part by the U.S. Department of Commerce under Grant BS123456.'' }
%\thanks{The next few paragraphs should contain the authors' current affiliations, including current address and e-mail. For example, F. A. Author is with the National Institute of Standards and Technology, Boulder, CO 80305 USA (e-mail: author@boulder.nist.gov).}
%\thanks{S. B. Author, Jr., was with Rice University, Houston, TX 77005 USA. He is now with the Department of Physics, Colorado State University, Fort Collins, CO 80523 USA (e-mail: author@lamar.colostate.edu).}
}

\markboth{Submitted to IEEE Signal Processing Letters. Last update \today}
{Shell \MakeLowercase{\textit{et al.}}: Bare Demo of IEEEtran.cls for IEEE Journals}
\maketitle
\begin{abstract}
We propose a soft gradient boosting framework for sequential regression that embeds a learnable linear feature transform within the boosting procedure. At each boosting iteration, we train a soft decision tree and learn a linear input feature transform $\mathbf{Q}$ together. This approach is particularly advantageous in high-dimensional, data-scarce scenarios, as it discovers the most relevant input representations while boosting. We demonstrate, using both synthetic and real-world datasets, that our method effectively and efficiently increases the performance by an end-to-end optimization of feature selection/transform and boosting while avoiding overfitting. We also extend our algorithm to differentiable non-linear transforms if overfitting is not a problem. To support reproducibility and future work, we share our code publicly.
\end{abstract}

\begin{IEEEkeywords}
Gradient boosting, sequential regression, soft decision trees%, domain adaptation
\end{IEEEkeywords}

\IEEEpeerreviewmaketitle

\section{Introduction}

\IEEEPARstart{I}{n} this letter, we study sequential (online) regression problem, which aims to regress the next value $y_{T+1}$ of a % univariate
series \( \{y_t\}_{t=1}^{T} \) (optionally conditioned on an exogenous features \(\{\mathbf x_t\}_{t=1}^{T}\)) given all information up to time \(T\).
\footnote{We denote all vectors by boldface lowercase letters (e.g.\ $\mathbf{u}$), and matrices by boldface capital letters (e.g.\ $\mathbf{U}$). For a vector $\mathbf{u}$ (or matrix $\mathbf{U}$), $\mathbf{u}^\top$ ($\mathbf{U}^\top$) is the transpose. The time index is indicated by a subscript; for instance, $\mathbf{u}_t$ is a vector at time $t$. $\mathbf{I}_d \in \mathbb{R}^{d \times d}$ is the $d$-dimensional identity matrix.}
This problem models 
%forms the basis of 
a wide range of applications, such as energy-load forecasting, IoT and traffic management \cite{traffic_spl, iot_pred_spl}, and has been studied extensively in the signal processing literature. 
%This setting underpins a wide range of forecasting tasks, from energy‑load prediction to inventory control, and has been studied extensively in the literature. %(\textcolor{blue}{references can be inserted here}).

Gradient‑boosted decision‑tree models
(e.g.\ XGBoost \cite{chen2016xgboost}, LightGBM \cite{ke2017lightgbm})
are the current state of the art as demonstrated in different real life competitions
\cite{m4conclusions,m5conclusions}.  Yet they struggle when the feature dimension
is large relative to the amount of available training data—a frequent
situation that occurs in practice (e.g.\ short sales histories paired with hundreds
of engineered covariates) \cite{data_scarcity_nature_reports,scarcity_spl}.  Exhaustive subset or interaction searches %(\textcolor{blue}{references can be inserted here})
of those covariates are usually restrictive, making them impractical in data‑scarce, high‑dimensional scenarios \cite{high-dim-feat-select, dim_reduct_spl}.

% \textbf{Contribution.}  
As a solution, we propose a boosting framework that overcomes this bottleneck by inserting a learnable linear feature transform between successive weak (base) learners in gradient-boosting.  
We concentrate on linear transforms to avoid overfitting; however, we also extend our algorithm to differentiable nonlinear transforms, with or without memory (e.g., neural networks, RNNs, and similar), or to standard boosting machines using hard decision trees (e.g., vanilla LightGBM, XGBoost, CatBoost).
Specifically, we
\begin{enumerate}
\item replace hard trees with soft (differentiable) decision
      trees, enabling end-to-end optimization,
      % radient computation w.r.t.\ feature transform parameters,
\item learn a transform matrix \(\mathbf Q\) to perform embedded feature selection/transformation at every boosting round, which is jointly optimized with the newly fitted tree,
\item demonstrate that this framework can be readily extended to nonlinear feature transformations, %Note that such further  additions can cause overfitting, defeating the purpose, hence we mainly concentrate on the basic approach. 
\item openly share our code for replicability of our results and encourage further research. \footnote{\url{https://github.com/huseyin-karaca/bsdtq}}
\end{enumerate}

% Algorithm \ref{alg:bsdtq} gives the full procedure; derivations are in Section~\ref{sec:prob_def}, and empirical results (synthetic \& real) are in Section~\ref{sec:simulations}.

%(\textcolor{blue}{Related Work (to be completed).})

%To place our method in a solid panorama, we first review the main ideas behind existing feature‐selection and boosted‐tree approaches. 

%General feature extraction algorithms—such as filter-based or wrapper-based feature selection methods \cite{quinlan1986induction, das2001filters, kohavi1997wrappers}—are usually computationally costly. They also tend to be too general-purpose and not directly suitable for integration into boosting frameworks. 
%Recent studies, such as \cite{emirhan_spl}, show that adding extra steps between weak learners in gradient boosting can improve performance. 
%( Kendime cite olduğu için kaldırdım ama akışaa iyi oturmuştu.) Moreover, this concept has been extended even to hierarchical learning scenarios, as shown in \cite{tumay_hierarchical_2025}.

There are also existing studies in literature that combine soft decision trees with gradient boosting \cite{feng2020softgradientboostingmachine}, %\cite{feng2020softgradientboostingmachine, delibalta_spl},
and others integrating feature selection concepts into boosting methods \cite{vision_gradient_feature_selection,xu2014gradient}. The method introduced in \cite{vision_gradient_feature_selection} rely only on features specific to certain applications such as HOG features used in computer vision, which limits its general applicability. % Bence buradan sonrası direkt kalksın daha iyi. Our approach differs by introducing a general linear feature transform. This makes our method flexible enough to handle many different types of transformations.
The other related work \cite{xu2014gradient} combines traditional gradient-boosted decision trees with feature selection. Their method works effectively when the dataset has many samples compared to the number of features. However, in many real-world problems, especially in financial forecasting, we may frequently face the opposite case: few data points relative to many features \cite{data-scarcity-forecasting}. %In these cases, their method is not suitable. \textcolor{blue}{cite needed}

%Therefore, there is a clear need for
Hence, we introduce an end-to-end optimized method for model training, especially for situations with many features and limited data. 
%As a solution, we propose a boosting framework that overcomes this need by inserting a learnable \emph{linear feature transform} between successive weak (base) learners in gradient-boosting.  Specifically, we
% \begin{enumerate}
% \item replace hard trees with \emph{soft} (differentiable) decision
%       trees, enabling gradient computation w.r.t.\ feature transform parameters;
% \item introduce a parameter matrix \(Q\) initialised as the identity and
%       updated via gradient descent at every boosting round, using the
%       gradient of the newly fitted tree.
% \end{enumerate}
Note that by this basic transformation, we avoid both overfitting and remedy the need for feature selection and/or transformation, where the nonlinear interactions are discovered by the boosted decision trees. 
%In this paper, we introduce such a method. Our algorithm, named \textsc{bsdt-q} (Boosted Soft Decision Trees with feature transform $\mathbf{Q}$, Algorithm~\ref{alg:bsdtq}), addresses these issues effectively. 
We demonstrate through experiments—both with synthetic and real-world data—that our method 
significantly increases the performance of boosted trees while introducing little computational overhead
%significantly outperform 
%can compete with
%current state-of-the-art models, 
in scenarios where data is scarce and feature space is large.

\section{Problem Definition}
%\vspace{-0.1cm}
\label{sec:prob_def}
% We denote all vectors by boldface lowercase letters (e.g.\ $\mathbf{u}$), and matrices by boldface capital letters (e.g.\ $\mathbf{U}$). For a vector $\mathbf{u}$ (or matrix $\mathbf{U}$), $\mathbf{u}^\top$ ($\mathbf{U}^\top$) is the transpose. The time index is indicated by a subscript; for instance, $\mathbf{u}_t$ is a vector at time $t$. $\mathbf{I}_d \in \mathbb{R}^{d \times d}$ is the $d$-dimensional identity matrix.
\subsection{Sequential Regression}
\noindent We sequentially observe $\{y_t\}_{t \ge 1},\, y_t \in \mathbb{R}$ along with the feature vectors $\{\mathbf{x}_t\}_{t \ge 1},\, \mathbf{x}_t \in \mathbb{R}^d$. 
Our goal is to predict $y_{t+1}$ at each $t$ in an online manner by constructing a function that depends only on the past information:
\[
\hat{y}_{t+1} \;=\; f_t(\{\ldots, y_{t-1}, y_t\}, \{\ldots, \mathbf{x}_{t-1}, \mathbf{x}_t\}).
\]
At each time step $t$, we output $\hat{y}_{t+1}$ and suffer a loss $\mathcal{L}\bigl(y_{t+1}, \hat{y}_{t+1}\bigr)$. 
The loss function $\mathcal{L}$ can be any differentiable function such as the squared error loss $(y_{t+1} - \hat{y}_{t+1})^2$. 
We aim to minimize the accumulated online error over time:
\(
\mathcal{L}_T \;=\;\sum_{t=1}^{T} \mathcal{L}\bigl(y_t,\; \hat{y}_t\bigr).
\)
Note that this main framework can be readily extended to different horizons, e.g., predicting $y_{t+2}, .., y_{t+n}$, either using direct or recursive approaches \cite{taieb2012recursive}.

\begin{algorithm}[t]
\small
%\SetInd{2em}{0.7em}
\SetInd{1em}{0.1em}
\SetAlgoNoLine
\caption{Boosted Soft Decision Trees with feature transform $\mathbf{Q}$ (\textsc{BSDT-Q})}
\label{alg:bsdtq}
\KwIn{$\mathbf{X}_{\text{train}}\in\mathbb{R}^{n\times d}$, targets $\mathbf{y}\in\mathbb{R}^{n}$, number of boosting rounds $K$}
\KwOut{$\{(\mathbf{Q}_{k},\text{SDT}_{k})\}_{k=1}^{K}$, where $\text{SDT}_k$ is the base Soft Decision Tree at  k'th step, $\mathbf{Q}_k$ is the feature transform matrix }
\BlankLine
\textbf{Initialize} ensemble prediction $\hat{\mathbf{y}}^{(0)}\leftarrow \mathbf{0}$, linear transform $\mathbf{Q}_{0}^{}\leftarrow \mathbf{I}_{d}$\; 
\For{$k\leftarrow 1$ \KwTo $K$}{
    %\textcolor{teal}{\#step‑level initialisation}\;
    Residual $\mathbf{r}_{k}\leftarrow \mathbf{y}-\hat{\mathbf{y}}_{k-1}$\;
    % Linear transform $Q_{k}^{(0)}\leftarrow I_{d}$\;
    Fit soft decision tree $\text{SDT}_{k}^{}$ on $\bigl(\mathbf{Q}_{k-1}^{}\mathbf{X}_{\text{train}},\,\mathbf{r}_{k}\bigr)$\;
    Find $\mathbf{Q}_{k}^{}$ (for a number of gradient-descent steps w.r.t.\ loss of $\text{SDT}_{k}^{}$)\;
    %Update residual $r^{(k,i+1)}\leftarrow r^{(k,0)}-\text{SDT}_{k}^{(i)}\!\bigl(Q_{k}^{(i)}X_{\text{train}}\bigr)$\;
\BlankLine
    %\textbf{// lock in best model for round $k$}\;
    % $\mathbf{Q}_{k}\leftarrow \mathbf{Q}_{k}^{(i)}$, \quad $\text{SDT}_{k}\leftarrow\text{SDT}_{k}^{(i)}$\;
    $\hat{\mathbf{y}}^{(k)}\leftarrow \hat{\mathbf{y}}^{(k-1)}+\text{SDT}_{k}\!\bigl(\mathbf{Q}_{k}\mathbf{X}_{\text{train}}\bigr)$\;
}
\Return{$\{(\mathbf{Q}_{k},\text{SDT}_{k})\}_{k=1}^{K}$}
\end{algorithm}

%\vspace{-0.1cm}
\subsection{Hard Decision Trees}
A hard decision tree (HDT) $f(\mathbf{z})$ can be written as a sum of leaf values multiplied by indicator functions
\[
f(\mathbf{z}) \;=\;
\sum_{n=1}^N\, \gamma_n \;\mathbf{1}\bigl(\mathbf{z} \in P_n\bigr),
\]

\noindent where $P_n$ denotes the region (leaf) in the input space associated with the $n$-th leaf, and $\mathbf{1}(\cdot)$ is the indicator function that returns 1 if $\mathbf{z}$ belongs to $P_n$ and 0 otherwise. Hard decision trees use indicator-based splits (e.g., if $z_j < \text{threshold}$, go left; else go right). These splits are not differentiable with respect to the input $\mathbf{z}$, hence there is no straightforward way to compute  
%$ \frac{\partial \,f\bigl(\mathbf{Q}\,\mathbf{x}\bigr)}{\partial \mathbf{Q}}$.
$\nabla_{\mathbf{Q}}f(\mathbf{Qx})$.
This makes gradient-based optimization of $\mathbf{Q}$ challenging. Nevertheless, in many real-world applications (including sequential regression), hard decision trees are crucial due to their interpretability and strong performance, especially when they are used in a gradient-boosting framework.

%\vspace{-0.1cm}
\subsection{Soft Decision Trees}
\label{subsec:sdt}
Soft decision trees (SDT) relax the hard splits of hard decision trees into differentiable “soft” gating functions, often using a logistic $\sigma(\cdot)$ at each node. Let $\ell$ index a leaf in the set of leaves $\boldsymbol{\ell}$, and let $\text{path}(\ell)$ denote the set of node indices along the path from the root to leaf $\ell$. Then each node $i$ outputs a probability $p_i(\mathbf{z}) = \sigma\bigl(\mathbf{w}_i^\top \mathbf{z} + b_i\bigr)$. Depending on whether the node “direction” $v_{i,\ell}$ is 1 (go right) or 0 (go left), the contribution to the final leaf value is $p_i(\mathbf{z})$ or $1 - p_i(\mathbf{z})$, respectively. Thus:
\begin{equation}
\label{eq:sdt0}
p_\ell^* \;=\;
\prod_{i \in \text{path}(\ell)} 
\bigl[p_i(\mathbf{z})^{\,v_{i,\ell}} \,\bigl(1 - p_i(\mathbf{z})\bigr)^{\,1 - v_{i,\ell}}\bigr].
\end{equation}
The overall tree output is then
\[
f(\mathbf{z}) \;=\;
\sum_{\ell \,\in\, \boldsymbol{\ell}}\;
p_\ell^*(\mathbf{z})\;\gamma_\ell,
\]
where $\gamma_\ell$ is the value assigned to the leaf $\ell$. Due to $\sigma(\cdot)$, $f(\mathbf{z})$ is differentiable with respect to \ $\mathbf{z}$, and hence with respect to $\mathbf{Q}$. We calculate a gradient-descent update to $\mathbf{Q}$, using differentiability of soft trees in Section \ref{sec:derivation}.

\subsection{Gradient Boosting Machines}
\label{sec:gbm}

Gradient Boosting Machines \cite{Friedman2001} build a strong predictor
by stage–wise accumulation of weak (base) learners.  
Given training data
\(\mathcal D=\{(\mathbf x_i,y_i)\}_{i=1}^N\)
and a differentiable loss \(\mathcal{L}\bigl(y,F(\mathbf x)\bigr)\),
the ensemble after \(m\) rounds is given by

\[
  F_{m}(\mathbf x)=F_{m-1}(\mathbf x)+\nu\,f_{m}(\mathbf x),
  \qquad 
  0<\nu\le 1,
  \label{eq:additive_gbm}
\]

\noindent where \(f_m\) is the new weak learner and \(\nu\) is a
shrinkage factor. At each round~\(m\) we learn the new weak learner \(f_m\) to be added to the ensemble as % \emph{pseudo‑residuals}

\[
  f_m=\arg\min_{f\in\mathcal H}
       \sum_{i=1}^N \bigl(g_{i,m}-f(\mathbf x_i)\bigr)^2
       \;+\;\Omega(f)
  \label{eq:fitting}
\]

\noindent where \(\mathcal H\) is the hypothesis space (e.g.\ hard
decision trees), \(\Omega\) is the regularizer. The $g_{i,m}$ represent the negative gradient of the loss function (with respect to the function values of the current ensemble):

\[
  g_{i,m}=
  \,\Bigl.
  -\frac{\partial\,\mathcal{L}(y_i,F)}
       {\partial F}
  \Bigr|_{F=F_{m-1}(\mathbf x_i)}
  \qquad i=1,\dots,N.
  \label{eq:pseudo_residual}
\]

%By doing so, at each round, we add the weak learner that focuses mostly on the current model's overlooked points. 
\noindent Thus, at each iteration we fit a new weak learner to the residual errors so that it focuses on the observations the ensemble currently predicts poorly.

\section{Learning an Efficient Feature Transform for Soft Decision Trees}
\label{sec:derivation}

%\subsection{Derivation of a Linear Transform}
When using a soft decision tree $f$ and an input transform $\mathbf{z} = \mathbf{Q}\,\mathbf{x}$, where $\mathbf{x} \in \mathbb{R}^{d_{in}}$, $\mathbf{z} \in \mathbb{R}^{d_{out}}$, $\mathbf{Q} \in \mathbb{R}^{d_{out} \times d_{in}}$, we can perform gradient-based updates of $\mathbf{Q}$ as 

\[
\mathbf{Q}_{k+1} \;\leftarrow\; \mathbf{Q}_k \;-\; \eta\,\nabla_{\mathbf{Q}_k}\Bigl[\mathcal{L}\bigl(y,\,f(\mathbf{Q}_k\,\mathbf{x})\bigr)\Bigr],
\]

\noindent where $\eta$ is the learning rate. Following the definitions given in Section \ref{subsec:sdt} and rewriting \eqref{eq:sdt0},  %sketch of the derivative steps is as follows: 
we have
%\subsection{\; Derivation of  \texorpdfstring{$\frac{\partial f(\mathbf{Qx})}{\partial \mathbf{Qx}}$}{df/dz}}

\[
p_{\ell}^*(\mathbf{z}) \;=\;
\prod_{i\in \text{path}(\ell)} 
\beta_{i,\ell}(\mathbf{z}),
\quad
\]
where
\[
\quad
\beta_{i,\ell}(\mathbf{z})
=
\begin{cases}
p_i(\mathbf{z}), & v_{i,\ell} = 1,\\
1 - p_i(\mathbf{z}), & v_{i,\ell} = 0,
\end{cases}
\]
and $p_i(\mathbf{z}) = \sigma(\mathbf{w}_i^\top \mathbf{z} + b_i)$. By the product rule,
\[
\frac{\partial p_{\ell}^*(\mathbf{z})}{\partial z_k}
=
p_{\ell}^*(\mathbf{z})\;
\sum_{m\in \text{path}(\ell)}
\frac{1}{\beta_{m,\ell}(\mathbf{z})} 
\,\frac{\partial \beta_{m,\ell}(\mathbf{z})}{\partial z_k}.
\]
Since $\beta_{m,\ell}(\mathbf{z})$ is either $p_m(\mathbf{z})$ or $1 - p_m(\mathbf{z})$, we get
\[
    \frac{1}{\beta_{m,\ell}(\mathbf{z})}\;\frac{\partial \beta_{m,\ell}(\mathbf{z})}{\partial z_k}
\;=\;
(v_{m,\ell} - p_m(\mathbf{z}))\; w_{m,k},
\]
where $\mathbf{w}_m = (w_{m,1},\ldots,w_{m,d_\mathrm{out}})^\top$ is the node's parameter vector. Summing over $m$ yields
\[
\frac{\partial p_{\ell}^*(\mathbf{z})}{\partial z_k}
=
p_{\ell}^*(\mathbf{z}) 
\;\sum_{m\in \text{path}(\ell)}\!\bigl[v_{m,\ell} - p_m(\mathbf{z})\bigr]\;w_{m,k}\,.
\]
Hence,
\begin{equation}
\label{eq:delfzdelzk}
\begin{split}
\frac{\partial f(\mathbf{z})}{\partial z_k}
&=
\sum_{\ell \,\in \boldsymbol{\ell}}
\gamma_{\ell}\;\frac{\partial p_{\ell}^*(\mathbf{z})}{\partial z_k}
\\&= 
\sum_{\ell \,\in \boldsymbol{\ell}}
\gamma_{\ell}\;
p_{\ell}^*(\mathbf{z})\;\sum_{m\in \text{path}(\ell)}\!\bigl[v_{m,\ell} - p_m(\mathbf{z})\bigr]\;w_{m,k}\, .
\end{split}
\end{equation}

%\subsubsection{\; Differentiating \texorpdfstring{$f(\mathbf{Q}\,\mathbf{x})$}{f(Qx)} w.r.t.\ \texorpdfstring{$\mathbf{Q}$}{Q}}
\noindent Let $\mathbf{z} = \mathbf{Q}\,\mathbf{x}$. Then for each entry $Q_{r,c}$,
\begin{equation}
\label{eq:chainrule}
\frac{\partial f(\mathbf{Q}\mathbf{x})}{\partial Q_{r,c}}
=
\sum_{k=1}^{d_{\mathrm{out}}}
\frac{\partial f(\mathbf{z})}{\partial z_k}
\;\frac{\partial z_k}{\partial Q_{r,c}}.
\end{equation}
However, $z_k = \sum_{j=1}^{d_{\mathrm{in}}} Q_{k,j}\,x_j$, thus $\partial z_k \, /\,
\partial Q_{r,c} = x_c$ if $k=r$ and 0 otherwise, making
\[
\frac{\partial f(\mathbf{Q}\mathbf{x})}{\partial Q_{r,c}}
=
\left(\frac{\partial f(\mathbf{z})}{\partial z_r}\right)\; x_c.
\]

\bigskip
\noindent
Finally, the elementwise result can be obtained as
\[
\frac{\partial f(\mathbf{Qx})}{\partial Q_{r,c}}=
x_c \sum_{\ell}
\gamma_{\ell}\;p_{\ell}^*(\mathbf z)\sum_{m\in\text{path}(\ell)}
\bigl[v_{m,\ell} - p_m(\mathbf z)\bigr]w_{m,r} \, .
\]
%Here, $w_{m,r}$ is the $r$-th component of node $m$'s weight vector $w_m$, and $\mathbf{z} = \mathbf{Q}\,\mathbf{x}$. 
Putting it all together yields the final expression for $\nabla_{\mathbf{Q}} f(\mathbf{Q}\mathbf{x})$. Then we can combine it with the loss $\mathcal{L}\bigl(y,f(\mathbf{Q}\mathbf{x})\bigr)$ to update $\mathbf{Q}$ via gradient descent. For example, in the case of $\mathcal{L}_2$ loss \(
 \tfrac12 (f(\mathbf{Qx}_i)\;-\;y_i)^{2},
\) the chain rule (vectorized over all samples) results
\[
\frac{\partial\mathcal{L}}{\partial Q_{r,c}}
\;=\;
([f(\mathbf{Qx})]\;-\;\mathbf{y})^{\!\top} \;
[\frac{\partial f(\mathbf{Qx})}{\partial Q_{r,c}}].
\]

%\noindent where $[a]$ represents vectorized a, entries beginning from $a_0$ to $a_N$. \textcolor{blue}{Burada daha doğru ve dikkatli olalım (matematiksel olarak).} 

\subsubsection*{\textbf{Remark 1} (Non-linear feature transforms)}
One can readily extend our linear transform to non-linear transforms.
The derivations leading up to ~\eqref{eq:delfzdelzk} rely only on the soft–decision-tree (SDT) structure; the specific form of the feature transform
\(
\phi\colon \mathbb{R}^{d_{in}}\!\to\!\mathbb{R}^{d_{out}},\;
\mathbf{x}\mapsto \mathbf{z}=\phi_{\Theta}(\mathbf x)
\)
does not appear.  
Consequently, we can replace the current linear map $\mathbf Q$ with any differentiable transform $\phi_{\Theta}(\cdot)$
%parameterized by \(\Theta\)
, such as NN's.  
Starting from ~\eqref{eq:chainrule} we 
%only have to change one chain-rule factor
need to change

\begin{equation}
  \frac{\partial f(\mathbf{z})}{\partial \theta}
  \;=\;
  \sum_{k=1}^{d_{out}}
    \underbrace{\frac{\partial f(\mathbf{z})}{\partial z_k}}_{\text{\tiny depends on SDT}}\;
    \underbrace{\frac{\partial z_k}{\partial\theta}}_{\text{\tiny depends on $\phi_{\Theta}$}}\;,
  \label{eq:full_chainrule}
\end{equation}
where $\theta$ any parameter $\in \Theta$, and only \(\partial z_k / \partial\theta\) depends on the chosen transform.
%Equation \eqref{eq:full_chainrule} is completely general; the term
%\(\partial z_k / \partial\theta\) is the only piece that depends on the chosen transform.
For example, if we use multi-layer perceptron  as \(\phi_{\Theta}\),
the explicit expression to place in  \eqref{eq:full_chainrule}  would be 

\[
\frac{\partial z_k}{\partial w^{(l)}_{ij}}
=
\frac{\partial z_k}{\partial a^{(l)}_i}\;
\frac{\partial a^{(l)}_i}{\partial w^{(l)}_{ij}}
=
\delta^{(l)}_i\;
h^{(l-1)}_j .
\]

% where element \(w^{(\ell)}_{ij}\) connecting unit \(j\) in layer \((\ell-1)\) to unit \(i\) in layer \(\ell\), $\mathbf{a}^{(\ell)}$ and $\mathbf h^{(\ell)}$ pre- and post-activation of layer $\ell$, and $\delta^{(l)}_i$ is the back-propagation vector for the single output coordinate  $\frac{\partial z_k}{\partial a^{(l)}_i} = \sigma'\!\bigl(a^{(l)}_k\bigr)$ only when $i=k $.

% % $\mathbf h^{(0)} = \mathbf x,\;
% % \mathbf{a}^{(\ell)} = \mathbf{W}^{(\ell)}\mathbf{h}^{(\ell-1)} + \mathbf{b}^{(\ell)},\;
% % \mathbf{h}^{(\ell)} = \sigma\!\bigl(\mathbf{a}^{(\ell)}\bigr),\;
% % \mathbf{z} = \mathbf{h}^{(L)}$

% Implementationwise, it is also easier, since all neural network libraries already records gradients, no manual derivation is needed in code.

\noindent Here \(w^{(l)}_{ij}\) denotes the weight that links neuron \(j\) in layer \(l-1\) to neuron \(i\) in layer \(l\).  Vectors \(\mathbf a^{(l)}\) and \(\mathbf h^{(l)}\) refer to the pre-activation and post-activation signals of layer~\(\ell\), respectively.  The single-output back-prop term is therefore  \(
\delta^{(l)}_i
\;=\;
\partial z_k \; / \;\partial a^{(l)}_i
\;=\;
\sigma'\!\bigl(a^{(l)}_k\bigr)\) only when  $i = k$.

%Modern deep-learning frameworks (e.g.\ PyTorch, TensorFlow, JAX) automatically construct the computational graph and supply these gradients via automatic differentiation, so no manual derivative calculation is needed in code.
%Equation~\eqref{xx} shows that, instead of a linear map, we could employ \emph{any} differentiable transform, for example a standard neural network. 
One could likewise plug in more expressive architectures such as LSTMs or CNNs in a similar manner. However, every additional layer increases the number of trainable parameters and, consequently, the risk of over-fitting, defeating the purpose of this letter. Hence we deliberately keep the transform lightweight.

\subsubsection*{\textbf{Remark 2} (Use of HDT as base learners)}
The non-differentiable splits of hard decision trees prevent us from computing
\(\partial\mathcal{L}/\partial\theta\) for the feature transform
\(\phi_{\theta}\).
To extend our framework to this case, we would have to do back-and-forth optimization between (i) fitting the tree on the current
features and (ii) adjusting \(\phi_{\theta}\) with the tree frozen, then repeating the cycle.  
This alternating scheme increases training time and may settle on a
configuration that is not jointly optimal for the tree and the transform as shown in our experiments.
\section{Simulations}
%%\vspace{-0.3cm}
\label{sec:simulations}

\begin{comment}
    
%- ablation study: Q optimization olmadan hem hard hem soft tree ile sequential prediction. (eğer boosting ya da hierarchical bir setting altında q optimization yapacaksak tabii ki karşılaştırma yapacağımız deneyler de o şekilde olacak.) base modelleri ve sota modelleri kullanabiliriz.

%- datasetler: 1) işe yaradığın göstermek için herkesin kullandığı m4, m5 gibi datasetlerde deney yapacağız.

2) önerdiğimiz algoritmanın nasıl işe yaradığını göstermek için sentetik datada deney yapacağız. bu sentetik datayı mesela sadece 5 feature'ın kullanıldığı bir tree'de üreteceğiz ve 25-30 tane gereksiz feature ekleyerek bizim kendi algoritmamızda test edeceğiz. Q matrisinin 25-30 tane gereksiz feature arasından bizim sentetik datayı üretirken kullandığımız gerçekten de anlamlı olan 5 feature'ı bulacak şekilde değiştiğini göstereceğiz. 
\end{comment}

\subsection{Synthetic Data Analysis} 
\begin{figure}
    \centering
    \includegraphics[width = 0.95\linewidth]{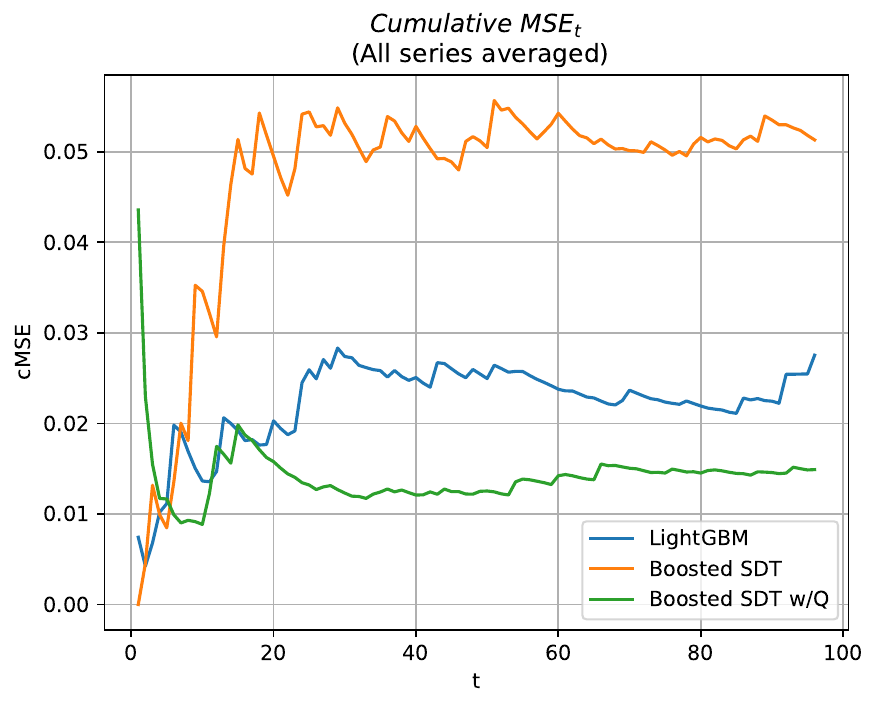}
    \caption{Comparative performance analysis on synthetic dataset (Cumulative average MSE) among three models: \textit{LightGBM}, classical \textit{Boosted Soft Decision Trees without Q transforms (BSDT)}, and \textit{Boosted Soft Decision Trees with Q transforms (BSDT-Q)}}
    \label{fig:synt_cumsum_mse}
    %\vspace{-0.25cm}
\end{figure}

To illustrate how our proposed method works, i.e., how the transform manages irrelevant features, we construct a dataset with only 5 truly predictive features, but artificially include 45 additional irrelevant ones. Namely, we have 50 univariate time--series, each covering \(196\) time steps.  
For every series \(s\in\{1,\dots,50\}\) and time \(t\in\{1,\dots,196\}\) we draw   five relevant exogenous signals \(x^{\text{rel}}_{s,t,i}\sim\mathcal N(0,1)\) and forty--five irrelevant signals of identical distribution.  
Independent weights \(w_i\sim\mathrm{U}(0.5,1.5)\) are sampled once and shared across series;  the target is generated by  
\begin{equation}
\label{eq:sentetik}
  y_{s,t}= \sum_{i=1}^{5} w_i\,x^{\text{rel}}_{s,t,i} \;+\; \varepsilon_{s,t},
  \qquad
  \varepsilon_{s,t}\sim\mathcal N(0,0.1^{2}).
\end{equation}
%The resulting long--format table contains the columns  \texttt{(unique\_id, ds, y, relevant\_1$\dots$5, irrelevant\_1$\dots$45)}.  
We use the first 100 samples for training and the last 96 for testing.
%Then the last \(t=96\) and the rest \(t=100\) yielded test and training sets, respectively.
%We then train our Boosted Soft Decision Trees with Q transforms (BSDT-Q) to see if it can effectively separate the irrelevant ones from the true input features. 
Then, we visualize the cumulative MSE
\footnote{For each sequence, denoted by \( y_t^{(i)} \), we apply all the algorithms on the test part and calculate the loss values 
\( l_t^{(i)}  \triangleq \left(y_t^{(i)} - \hat{y}_t^{(i)}\right)^2 \) on test days for each algorithm. 
We then take the average over all \( i \) to remove the effect of individual sequences and compute
the mean squared error as 
\(
MSE_t = \frac{1}{N} \sum_{i=1}^{N} l_t^{(i)}
\)
for each time instant. To further smooth the results over time, we also compute a cumulative time-averaged version:
\(
cMSE_t= \frac{1}{t} \sum_{j=1}^{t} MSE_t.
\)
}
over time 
%and the $\mathbf{Q}$  matrix of the final boosting iteration as 
%$|\mathbf{Q}-\mathbf{I}_d|$ 
as in the Fig. \ref{fig:synt_cumsum_mse}.
%and Fig. \ref{fig:fig2_Q}, respectively. 

% As expected, our method achieved visibly lower error than other methods and 
% %Due to the small learning rate in gradient descent steps, the updated entries of \( \mathbf{Q} \) become significantly smaller compared to the diagonal entries, which initially equal 1. Thus, subtracting the identity matrix \( \mathbf{I}_d \) is essential to clearly observe meaningful updates. 
% %As expected, the plot shows 
% clear enhancing patterns on entries corresponding to the relevant features observed.
% %only, i.e., the first five features, proving effectiveness of our proposed algorithm.
% Furthermore, we also observed the rows of our $\mathbf{Q}$ matrix converged to the original weights $w_i$ after normalization.

% \begin{figure}[h]
%     \centering
%     \includegraphics[width=\linewidth]{synt_q_5_by_50.pdf}
%     \caption{Visualization of the $\mathbf{Q}$ matrix for the synthetic simulation. Output dimension is selected as 5 in this case.}
%     \label{fig:fig2_Q}
% \end{figure}
   
\begin{comment}
\subsection{Notes}
\begin{itemize}
\item Across all experiments, we  report both the predictive performance and the computational cost, emphasizing the trade-off between model accuracy and the added complexity of learning a transform.

\item Although the primary experiments focus on single or boosting-based decision trees, our methodology could naturally extend to hierarchical or other advanced tree designs, such as heterogeneous boosting.
\end{itemize}
\end{comment}

To evaluate the feature selection capability of our proposed algorithm, we again generated a synthetic dataset using a linear relationship as defined in \eqref{eq:sentetik}, with 3 relevant and 2 irrelevant features. The ground-truth weights were selected randomly at first and kept fixed. We train our algorithm independently 100 times with different initial conditions. The resulting mean and standard deviation of the learned weights (entries of the $\mathbf Q$) are presented in Table \ref{tab:values}.

As can be seen from Table \ref{tab:values}, the learned weights converge to the true weights, and particularly, the irrelevant features consistently approach to zero, demonstrating the robustness and consistency of our method.

\begin{table}
\label{tab:values}
\centering
\caption{Mean and standard deviation of the learned weights across 100 independent runs. 
%The ground‐truth weights are set to \(w=[10,7,4,0,0]\). Irrelevant features (\(w_4,w_5\)) consistently converge to values close to zero.
}
\begin{tabular}{c c c c}
\toprule
\textbf{Feature}  & \textbf{Actual value} & \textbf{Mean} & \textbf{Standard Deviation} \\
\midrule
\(w_1\) &  0.476 &  0.489 & 0.0016 \\
\(w_2\) &  0.333 &  0.309 & 0.0014 \\
\(w_3\) &  0.190 &  0.175 & 0.0009 \\
\(w_4\) &  0.000 &  0.023 & 0.0021 \\
\(w_5\) &  0.000 &  0.004 & 0.0007 \\
\bottomrule
\end{tabular}
%\vspace{-0.25cm}
\end{table}

\subsection{Testing on Real--World Time Series}%
\label{sec:real_data}

We next evaluate our algorithm over 
%To demonstrate that our method is not confined to synthetic benchmarks, we evaluate it on 
three well-known publicly--available, real–life datasets that represent
diverse application domains and sampling frequencies:

\begin{itemize}
  \item Exchange (daily).  
        8 national exchange rates against the USD observed
        between 1990 – 2016, covering macro–economic and geopolitical
        events over more than two decades \cite{lai2018modeling}.
  \item ETTh2 (hourly).  
        Electricity transformer data collected in Eastern China (July 2016 – July 2018), including oil temperature and several load variables\,\cite{zhou2021informer}.
  \item Weather (15-minute).  
        Twenty-one meteorological measurements from the Max-Planck
        Biogeochemistry station in Jena (calendar year 2020).  
        Although the raw file contains many channels, we forecast only
        the ambient air temperature ($^{\circ} C$) time series
        %,treating the remaining variables as exogenous features\,
        \cite{wu2021autoformer}.
    \item M4 (quarterly). A publicly-available collection of real-world time series from the M4 forecasting competition, drawn from domains such as demographics, finance, industry and tourism \cite{makridakis2020m4}. We select the quarterly series from the seven sampling frequencies since these are most data scarce series in the competition.
\end{itemize}

%These datasets collectively exercise our regressor on \emph{daily}, \emph{hourly}, and \emph{sub-hourly} cadences, confirming that the model remains stable across markedly different temporal resolutions.

%\paragraph{Feature construction.}
During preprocessing we first scale each target to the
interval~$\left[-1,1\right]$ by a per-series min–max transform.  We then
generate $\sim$ 50 lagging and rolling
statistics as features.
%whose window lengths are expressed in native time steps (e.g.\ \texttt{lag\_7}, \texttt{roll\_mean\_48}, \texttt{roll\_std\_168} for the hourly ETTh2 data). 
%\paragraph{Train–test protocol.}
For every time series we reserve the last 96 time steps as a hold-out test window and draw a random subset of 100 time steps from the remaining observations for training.
This procedure purposefully creates a
data scarce, high dimensional setting that our algorithm is designed to
handle.
%\footnote{%Preliminary experiments with larger training windows did not materially change the ranking of methods but diluted the intended small-sample regime, so we keep the budget at 100~points throughout.}
%\paragraph{Results.}

\begin{figure}[t]
    \centering
    \includegraphics[width = 0.95\linewidth]{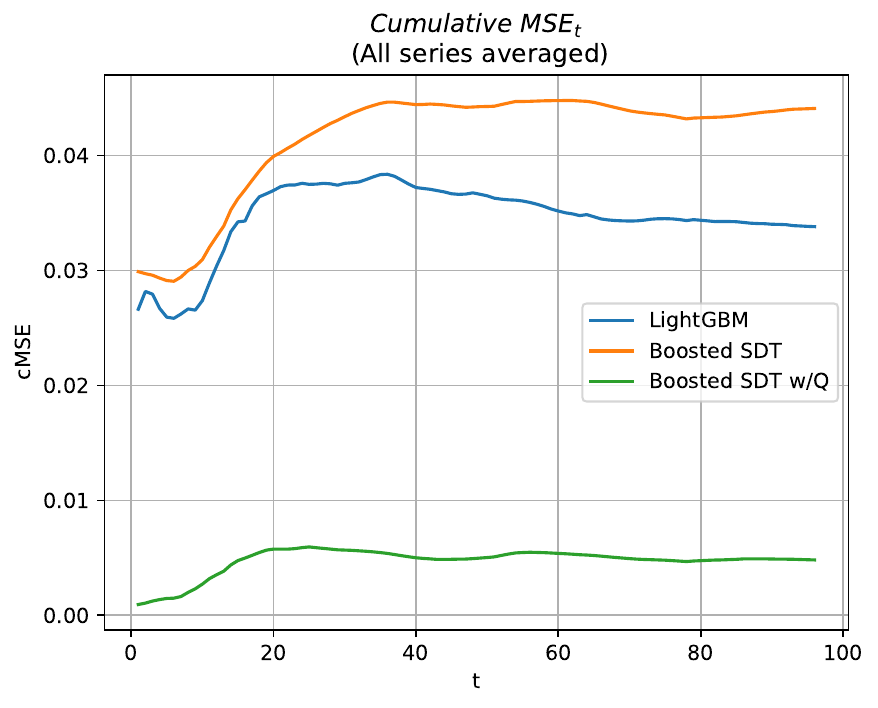}
    \caption{Comparative performance analysis on Exchange dataset \cite{lai2018modeling} (Cumulative average MSE) among three models: \textit{LightGBM}, classical \textit{Boosted Soft Decision Trees without Q transforms (BSDT)}, and \textit{Boosted Soft Decision Trees with Q transforms (BSDT-Q)}.}
    \label{fig:exchange_cumsum_mse}
    %\vspace{-0.25cm}
\end{figure}

All models are hyperparameter-tuned via grid search and then evaluated on a separate test set. We observe that our algorithm significantly outperforms the baseline BSDT and the LightGBM algorithms. Fig.~\ref{fig:exchange_cumsum_mse} shows the cMSE curve for the Exchange data, and Table~\ref{tab:mse_all} shows average MSE over test days on all other datasets.

%Quantitative results for all datasets are summarized in Table~\ref{tab:mse_real}. It is observed that our algorithm significantly outperform its baseline model BSDT and 

\begin{table}
  \centering
  \caption{Test-set mean--squared error (MSE) for all datasets.}
  \label{tab:mse_all}
  \setlength{\tabcolsep}{6pt}
  \begin{tabular}{lccc}
    \toprule
    \textbf{Dataset} & \textbf{BSDT-Q} & \textbf{BSDT} & \textbf{LightGBM}\\
    \midrule
    Exchange & \textbf{0.0048}  &0.0441 & 0.0338\\
    ETTh2    & \textbf{0.0040}  & 0.0084 & 0.0098 \\
    Weather  & \textbf{0.0012} & 0.0115 & 0.0022 \\
    M4 (quarterly) & \textbf{0.0124} & 0.0300 & 0.0291 \\
    \midrule
    Synthetic & \textbf{0.0143} & 0.0510 & 0.0275 \\
    \bottomrule
  \end{tabular}
  %\vspace{-0.5cm}
\end{table}

\section{Conclusion}
\label{sec:conclusion}

% In this letter, we proposed a \emph{soft gradient boosting} framework that incorporates a learnable linear feature transform. Specifically, each boosting round consists fitting a new soft decision tree on the inputs and updating a transform matrix $\mathbf{Q}$ via gradient descent on the overall loss function. We demonstrated that this approach enhances the feature selection and representation capability of boosted trees, leading to improved performance in challenging high-dimensional, data-scarce scenarios. We derived all the required update rules and showed how the same approach can be extended to other differentiable transforms. Through experiments on both synthetic and real-world datasets, our method matched or exceeded the performance of standard boosted-tree models in data-scarce, high-dimensional settings. To support reproducibility and further work, we have made our implementation publicly available.

In this letter, we introduce a soft gradient boosting framework for sequential regression that integrates a learnable linear feature transform into boosting. We fit a soft decision tree on the current inputs and then learn a transform matrix \(\mathbf{Q}\) in an end-to-end manner, allowing feature selection and model fitting together at each boosting iteration.
%We derived all the required update rules and showed how the same approach can be extended to other differentiable transforms. 
We also show how the same approach can be extended to other differentiable transforms and hard decision trees.
Through experiments on both synthetic and real-world datasets,
we show that our algorithm notably increases the performance of boosted trees while adding minimal computational overhead
%our method matched or exceeded the performance of standard boosted-tree models
in data-scarce, high-dimensional settings. To support reproducibility and further work, we make our implementation publicly available.

\bibliographystyle{IEEEtran}
\bibliography{ref}

\end{document}